\documentclass{article} 
\usepackage{iclr2026_conference, times}
\usepackage{times}

\usepackage{amsmath,amsfonts,bm}

\def\eqref#1{equation~\ref{#1}}

\def\1{\bm{1}}

\DeclareMathAlphabet{\mathsfit}{\encodingdefault}{\sfdefault}{m}{sl}
\SetMathAlphabet{\mathsfit}{bold}{\encodingdefault}{\sfdefault}{bx}{n}

\usepackage{hyperref}
\hypersetup{
    unicode,                    
    pdfborder       = {0 0 0},  
    bookmarksdepth  = subsection,
    bookmarksopen   = true,     
    breaklinks      = true,
    colorlinks      = true,
    linkcolor       = citecolor,
    citecolor       = citecolor,
    urlcolor        = citecolor,
}
\usepackage{url}

\usepackage[table,xcdraw,dvipsnames]{xcolor}
\definecolor{citecolor}{HTML}{1F3B4D}

\usepackage[final]{microtype}
\usepackage{bm}
\usepackage{mathtools}

\usepackage{enumitem}
\usepackage{wrapfig}

\usepackage{caption}
\usepackage{subcaption}
\usepackage{graphicx}
\usepackage{dblfloatfix}
\usepackage{array}
\usepackage{rotating} 
\usepackage{float} 

\usepackage{tikz}

\usepackage{tabularx}

\definecolor{myorange}{RGB}{255,165,0}
\definecolor{myblue}{RGB}{0,0,128}
\definecolor{mygreen}{RGB}{0,128,0}
\definecolor{myred}{RGB}{255,10,10}
\definecolor{myviolet}{RGB}{138,43,226}

\definecolor{observed}{RGB}{170,170,170}
\definecolor{inference}{RGB}{255,91,89}
\definecolor{gptfouro}{HTML}{08306B}
\definecolor{claude}{HTML}{66C2A5}
\definecolor{gemini}{HTML}{A6D854}
\definecolor{humans}{HTML}{E7298A}
\definecolor{gptthreefive}{HTML}{6BAED6}
\definecolor{CAblue}{RGB}{15,96,125}
\definecolor{CAlightblue}{RGB}{58,160,171}
\definecolor{CAgreen}{RGB}{44,160,44}
\definecolor{CAred}{RGB}{214,39,40}
\definecolor{CApurple}{RGB}{148,103,189}
\definecolor{CAblueBg}{RGB}{5,60,90}
\definecolor{CAblueText}{RGB}{5,60,90} 

\let\oldemph\emph
\renewcommand{\emph}[1]{\textcolor{CAlightblue}{\oldemph{#1}}}

\usepackage{amsthm}
\usepackage{thmtools}
\usepackage{thm-restate}
\usepackage{mdframed}
\declaretheoremstyle[
    numberwithin=section,
    headfont=\normalfont\bfseries\sffamily \itshape\color{CAblue!80!black},
    notefont=\normalfont,
    bodyfont=\normalfont,
    headpunct={\vspace{0.0\topsep}},
    postheadspace={0.0em}, 
    mdframed={
            backgroundcolor=CAlightblue!10,
            linecolor=CAblue,
            rightline=false,
            topline=false,
            bottomline=false,
            linewidth=0.5em,
            splittopskip=\baselineskip,
            skipabove=1em,
            innertopmargin=0.8em,
            innerbottommargin=0.4em,
        },
]{tldrstyle}

\declaretheorem[style=tldrstyle,numbered=no,name=]{tldr*}

\usepackage{tocloft}
\newlistof{tldrs}{tldr}{\listtldrname}
\newcommand{\listtldrname}{List of TL;DRs}
\AtBeginEnvironment{tldr}{%
  \addcontentsline{tldr}{tldr}{\protect\numberline{\thetldr}TL;DR}
}

\DeclareRobustCommand{\colordot}[1]{\begin{tikzpicture}[baseline=(a.south)]
    \node[circle, scale=0.75,color=black, fill=#1] (a) {};
\end{tikzpicture}}

\DeclareRobustCommand{\dashedcircle}{%
    \begin{tikzpicture}[baseline=(a.south)]
        \node[circle, scale=0.75, draw=black, dashed, dash pattern=on 1pt off 1pt, fill=white] (a) {};
    \end{tikzpicture}%
}

\DeclareRobustCommand{\colordotcircum}[1]{\begin{tikzpicture}[baseline=(a.south)]
    \node[circle, scale=0.75,color=black, fill=white] (a) {};
\end{tikzpicture}}

\definecolor{CAblue}{RGB}{10,80,110} 

\definecolor{CAlightblue}{RGB}{58,160,171}
\definecolor{CAgreen}{RGB}{44,160,44}
\definecolor{CAred}{RGB}{214,39,40}
\definecolor{CApurple}{RGB}{148,103,189}

\DeclareRobustCommand{\cotcircle}{%
    \begin{tikzpicture}
        \draw[fill=CAblue, thick] (0,0) circle (0.35em);
    \end{tikzpicture}%
}

\DeclareRobustCommand{\numericcircle}{%
    \begin{tikzpicture}
        \draw[fill=CAlightblue, thick] (0,0) circle (0.35em);
    \end{tikzpicture}%
}

\newlength{\bibleftmargin}

\usepackage[
    capitalize,
    nameinlink,
]{cleveref}

\title{Do LLMs Share Human-Like Biases? Causal Reasoning Under Prior Knowledge, Irrelevant Context, and Varying Compute Budgets}

\author{Hanna M. Dettki \\
New York University\\
University of Tübingen\\
\texttt{\href{mailto:hmd8142@nyu.edu}{hmd8142@nyu.edu}} \\
\And
Charley M. Wu \\
Technical University Darmstadt\\
\And
 Bob Rehder  \\
New York University \\
}

\begin{document}
\iclrfinalcopy
\maketitle

\begin{abstract}
Large language models (LLMs) are increasingly used in domains where causal reasoning matters, yet it remains unclear whether their judgments reflect normative causal computation, human-like shortcuts, or brittle pattern matching.
We benchmark 20+ LLMs against a matched human baseline on 11 causal judgment tasks formalized by a collider structure ($C_1 \!\rightarrow\! E\! \leftarrow \!C_2$).
We find that a small interpretable model compresses LLMs' causal judgments well and that
most LLMs exhibit more rule-like reasoning strategies than humans who seem to account for unmentioned latent factors in their probability judgments.
Furthermore, most LLMs do not mirror the characteristic human collider biases of weak explaining away and Markov violations.
We probe LLMs' causal judgment robustness  under (i) semantic abstraction  and (ii) prompt overloading (injecting irrelevant text), and find that  chain-of-thought (CoT) increases robustness for many LLMs.
Together, this divergence suggests LLMs can complement humans when known biases are undesirable, but their rule-like reasoning may break down when uncertainty is intrinsic, as is in many real-world scenarios. This highlights the need to characterize LLM reasoning strategies for safe, effective deployment.
\end{abstract}

\section{Introduction}
Human causal judgment is powerful but brittle: people routinely deviate from normative inference under limited attention, fatigue, and contextual framing \citep{Danziger2011,Venkatraman2007,Barnes2015,Wittbrodt2018}. This motivates AI-assisted decision-making in high-stakes settings where consistent causal analysis could complement human limitations.
This is especially important since large language models (LLMs) are increasingly deployed in domains where causal understanding matters (e.g., legal and medical decision support) \citep{Dahl2024LargeLegal,riedemann2024path}.
Successful deployment therefore depends not only upon whether LLMs can generate correct output in known contexts via \textit{pattern matching / associative reasoning}, but crucially on whether they can apply \textit{causal reasoning strategies} allowing them to generalize to new contexts.

A core challenge is that in the real world, many causal judgment settings are ill-defined (e.g., no exact base rates of events are known) and hence humans rely on heuristics and prior assumptions, often introducing biases. 
Before deploying LLMs in high-stakes settings, it is hence paramount to not only compare LLM judgments to humans and investigate whether they reproduce certain  human biases, provided that LLMs have been trained on largely human generated data, but also whether LLMs provide sensible causal judgments (e.g., that the probability of a variable increases as more evidence arises).

\textbf{Related Work.}
Recent work evaluates LLMs on causal benchmarks with fully specified generative models, such as CLADDER \citep{jin2023cladder}, where queries have a single numeric ground truth given explicit base rates and link strengths. Such settings primarily test whether models can \textit{apply} statistical rules, but they are less informative about an agent's implicit world model (e.g., assumed priors or causal strengths) when these quantities are underspecified, as is often the case in real-world settings.
We therefore adopt a classic paradigm from human causal cognition \citep{rehder2003causal,rehder2014independence,bender2020causal,waldmann2006beyond} that deliberately leaves causal tasks underspecified, enabling a direct probe of the assumptions agents bring to the task; it also avoids requiring formal training in causal statistics, making human baselines more straightforward to collect.

Our study connects to the small body of work directly comparing humans and LLMs on the \textit{same} reasoning tasks \citep{gandhi2024humanlike,lampinen2024language,keshmirian2024biased}. It also relates to evidence that LLM judgments can shift under irrelevant context or superficial content changes \citep{pmlr-v202-shi23a,mirzadeh2024gsm}, and that Chain-of-Thought (CoT) can sometimes mitigate such fragility---addressing concerns that apparent causal reasoning may reflect learned correlations rather than causal understanding \citep{willig2023causal}. Similarly, \citet{zhu2024incoherent} study probabilistic \textit{coherence} in LLM judgments via repeated elicitation and violations of probability identities, using Bayesian cognitive models to interpret systematic deviations; our work instead examines the coherence and robustness of qualitative reasoning signatures across different causal queries and content manipulations.

Within collider structures, humans reliably exhibit systematic departures from normative qualitative patterns---including \textit{too little} explaining away \citep{fernbach2013cognitive,rottman2014reasoning} and Markov violations in which beliefs about one cause depend on information about an alternative cause \citep{ali2011mental,mayrhofer2016sufficiency,park2013mechanistic,rehder2017failures}. Whether and when LLMs reproduce these human-like biases, particularly given training on human-generated text, remains unclear.
Finally, while \citet{dettki2025large} tested against the same human baseline but evaluated only four LLMs, substantially broader evaluations and systematic robustness analyses remain limited, motivating the present work.

\textbf{Contributions.}
We pursue the question of how LLMs' causal reasoning compares to that of humans  by evaluating $20\!+\!$ LLMs on a causal inference benchmark derived from classic collider-graph judgments with a human baseline \citep{rehder2017failures}. We are interested in three core questions:
\textit{(Human-comparison)} do models reproduce human-like response patterns and biases?
\textit{(Mechanism)} can their responses be compactly explained by a normative causal Bayesian network (CBN) model?
\textit{(Robustness)} do these behaviors persist under content manipulations that reduce reliance on memorized world knowledge as well as in the face of inserted irrelevant information?
We test the above both with direct prompting \numericcircle{} where we ask for a single probability judgment as a response and with chain-of-thought (CoT) \cotcircle{} \citep{wei2022chain} where we ask LLMs to first reason step-by-step. Together, this yields a cognitively-grounded account of \textit{how} different LLMs reason causally---often more rule-based than humans and mostly not replicating human biases, with systematic shifts under CoT prompting.
We additionally release an LLM-friendly version of the  causal inference benchmark (including the human baseline) and
 a software package \href{https://github.com/hmd101/causAIign}{\textsc{causAIign}}  that supports structure-matched custom prompts and content manipulations.

\section{Methods}
\begin{wrapfigure}[19]{r}{0.6\textwidth}
    \centering
    \begin{subfigure}[b]{0.32\linewidth}
        \centering
        \includegraphics[width=\linewidth]{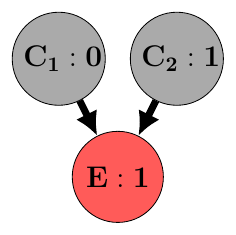}
        \caption{Task 2}
        \label{fig:sub_a}
    \end{subfigure}\hfill
    \begin{subfigure}[b]{0.32\linewidth}
        \centering
        \includegraphics[width=\linewidth]{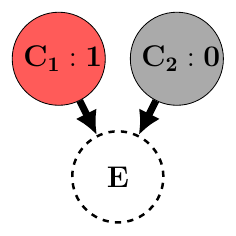}
        \caption{Task 5}
        \label{fig:sub_b}
    \end{subfigure}\hfill
    \begin{subfigure}[b]{0.33\linewidth}
        \centering
        \includegraphics[width=\linewidth]{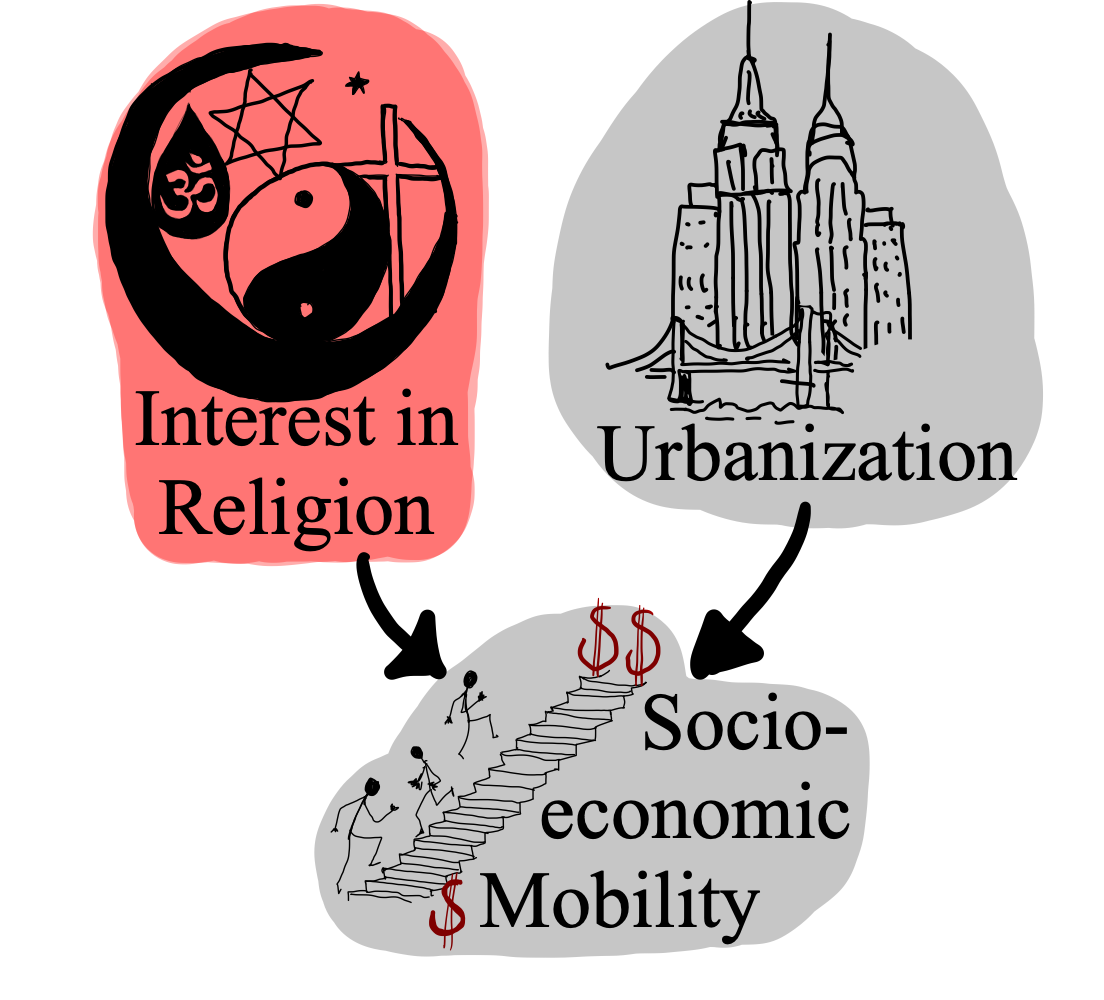}
        \caption{Domain Sociology}
        \label{fig:sub_c}
    \end{subfigure}
    \caption{\small \Cref{fig:sub_a} and \Cref{fig:sub_b} represent two out of 11 conditional probability (inference) queries. The colors of the nodes represent the three possible states the variables that the nodes of the graph can be in:
    \colordot{observed} $\!\to$\! observed node $\!\in\! \{0,1\}$,
    \colordot{inference} $\!\to\!$ latent (query node), \dashedcircle{} $\!\to\!$ no information. \Cref{fig:sub_c} shows an example of one of the three domains (Sociology) used in the RW17 conditions that ``verbalize'' the graph, where each node is represented by a variable in the domain that takes on binary states, e.g.,  $C_2$ being on (1) / off (0) translating to ``high /low urbanization''  in the sociology domain.
    }
    \label{fig:task-introx}
\end{wrapfigure}

\textbf{Benchmark tasks and human baseline. }
We build on causal inference tasks from \cite{rehder2017failures} (RW17; Experiment~1, Model-Only condition; $N{=}48$ NYU undergraduates), which elicit probability judgments in a collider graph $C_1 \!\rightarrow\! E \!\leftarrow \!C_2$. Variables are binary, causes are asserted as independent, and participants report probabilities (scale 0--100) for a target event (``off''/0 or ``on''/1) given observed values of the other variables. \cite{rehder2017failures} test 11 inference tasks (I--XI) corresponding to distinct conditional-probability queries on this collider---e.g., predictive queries such as $p(E{=}1 \mid C_1{=}0, C_2{=}1)$ in \Cref{fig:sub_a} and \Cref{fig:sub_pred_inference}---embedded in three different cover stories (sociology, weather, economy).
 Critically, because priors over variables and probabilities such as $p(E\!\mid\!C_1,\! C_2)$ are not specified, there is no ground-truth ``correct'' probability; the benchmark is therefore designed to assess \textit{reasoning strategies} of causal judgment (e.g., how consistent agents reason across tasks, or how rule-following they are) and \textit{qualitative patterns}, e.g.,  assessing the level of  explaining away and Markov-(non) compliance and compare them across agents and to the human baseline.

\textbf{LLM prompting and experimental conditions. }
For LLMs, we convert the RW17 training and test-phase information given to humans into a single textual prompt that includes (i) a domain introduction, (ii) causal mechanism statements for both causes, (iii) the observed evidence, and (iv) an inference question requesting a single numeric probability judgment  $\!\in \![0,100]$. We evaluate two prompting strategies: \textit{direct} prompting (single numeric answer) and \textit{chain-of-thought} (CoT) prompting (``think step by step'' before the numeric answer). We set temperature to $0.0$ to minimize sampling variance; for models that expose an explicit reasoning-budget control, we use it as provided by the API.

We test robustness via two content manipulations crossed with prompting, yielding $2\! \times \!2\! \times\! 2 \!= \!8$ experimental conditions:
(i) \textit{Prior knowledge}: original RW17 cover stories vs.\! \textit{abstract} variants in which RW17 semantically meaningful variable names are replaced by randomly generated strings, reducing reliance on real-world knowledge;
(ii) \textit{Information load}: original prompts vs.\! \textit{overloaded} prompts that append irrelevant text to reduce the signal-to-noise ratio (distraction);
(iii) \textit{Prompting}: direct vs.\! chain-of-thought (CoT).
Across all conditions, the dependent variable is the agent's probability judgment (0--100) for each of the 11 conditional probability queries that we refer to as inference tasks (cf.\! \Cref{fig:task-introx}).

\textbf{Causal stories and prompt structure}
Below is an example prompt including its prompt scaffolding  that conveys the causal story and prompt structure in a condensed form which all tasks regardless of their experimental condition shared. Here we present the \textsc{abstract} version where initially meaningful variables e.g., from the economy domain in \citet{rehder2017failures} are replaced with meaningless placeholders consisting of a random sequence of ten characters.
The \emph{blue text} illustrates injected noise that was  part of the \textit{overloaded} experimental conditions abbreviated with the suffix ``\textsc{-Over}''.
The injected noise was either neutral coming from the lorem ipsum vocabulary or meaningful but meaningless to the task coming from the original prompts in \citet{rehder2017failures} such that for example in the \textsc{RW17-Over} condition a prompt from the weather domain would contain a distracting sentence from the sociology domain.\footnote{The \textbf{bold} text is there to help the reader understand the structure of the prompt. Only the italicized text is part of the actual prompt given to LLMs.}

 \begin{small}
	- \textbf{Domain introduction:}
	\textit{In abstract reasoning studies, researchers examine relationships between symbolic variables \texttt{u8jzPde0Ig}, \texttt{xLd6GncfBA}, and \texttt{epfJBd0Kh8}.}\\
	- \textbf{Variable descriptions (\emph{overloaded}):}\\
	- \textbf{$C_1$ (\texttt{u8jzPde0Ig}).} \textit{Some systems have high \texttt{u8jzPde0Ig}. Others have low \texttt{u8jzPde0Ig}.}
\emph{Interest rates are the rates banks charge to loan money.}\\
	- \textbf{$C_2$ (\texttt{xLd6GncfBA}).} \textit{Some systems have weak \texttt{xLd6GncfBA}. Others have strong \texttt{xLd6GncfBA}.}\\
	- \textbf{$E$ (\texttt{epfJBd0Kh8}).} \textit{Some systems have weak \texttt{epfJBd0Kh8}. Others have powerful \texttt{epfJBd0Kh8}.}\\
	- \textbf{Causal relationships}
	\textit{Here are the causal relationships:}
\begin{itemize}
      \item \textit{High \texttt{u8jzPde0Ig} causes weak \texttt{epfJBd0Kh8}.}
      \emph{The good economic times produced by \dots .}
      \item \textit{Weak \texttt{xLd6GncfBA} causes weak \texttt{epfJBd0Kh8}.}
      \emph{Lorem ipsum dolor sit amet \dots .}
\end{itemize}
\textit{Both high \texttt{u8jzPde0Ig} and weak \texttt{xLd6GncfBA} can independently bring about weak \texttt{epfJBd0Kh8}.}\\
	- \textbf{Observations:}
	\textit{You are currently observing: weak \texttt{epfJBd0Kh8} and high \texttt{u8jzPde0Ig}.}\\
	- \textbf{Task instruction}
	\textit{Your task is to estimate how likely it is that weak \texttt{xLd6GncfBA} is present on a scale from 0 to 100, given the observations and causal relationships described. \dots .
}
 \end{small}


\section{Results}

We organize results by the following questions:
(Q1) \emph{Are Humans and LLMs aligned} in their responses and do agents provide sensible judgments;
(Q2) \emph{Do agents reason normatively} as measured via small, interpretable causal models.
(Q3) \emph{how rule-following are agents},
(Q4) \emph{are LLMs biased like humans}; and
(Q5) \emph{whether LLMs reason robustly} across content manipulations.

 \textbf{No significant domain differences.}
Within each agent, probability-judgment distributions did not differ across  domains (Kruskal--Wallis, $k{=}3$, $df{=}2$; Benjamini--Hochberg across agents: GPT-3.5-turbo $p_{\text{FDR-BH}}{=}.146$; all others $p_{\text{FDR-BH}}\ge .742$), so we pool the domains in all remaining analyses.
\subsection{Q1: Are Humans and LLMs aligned and do agents provide sensible judgements on our tasks?}

\vspace{1em}
\begin{tldr*}
  Both LLMs and humans provide sensible judgments on our tasks and judge the effect as more probable the more causes are present. 
  For less aligned models in the direct prompting condition, chain-of-thought prompting improves alignment with human judgments up to a ceiling effects at 0.85 (Spearman $\rho$).
\end{tldr*}

\begin{figure}
    \centering
    \begin{subfigure}[b]{0.32\linewidth}
        \centering
        \includegraphics[width=\linewidth]{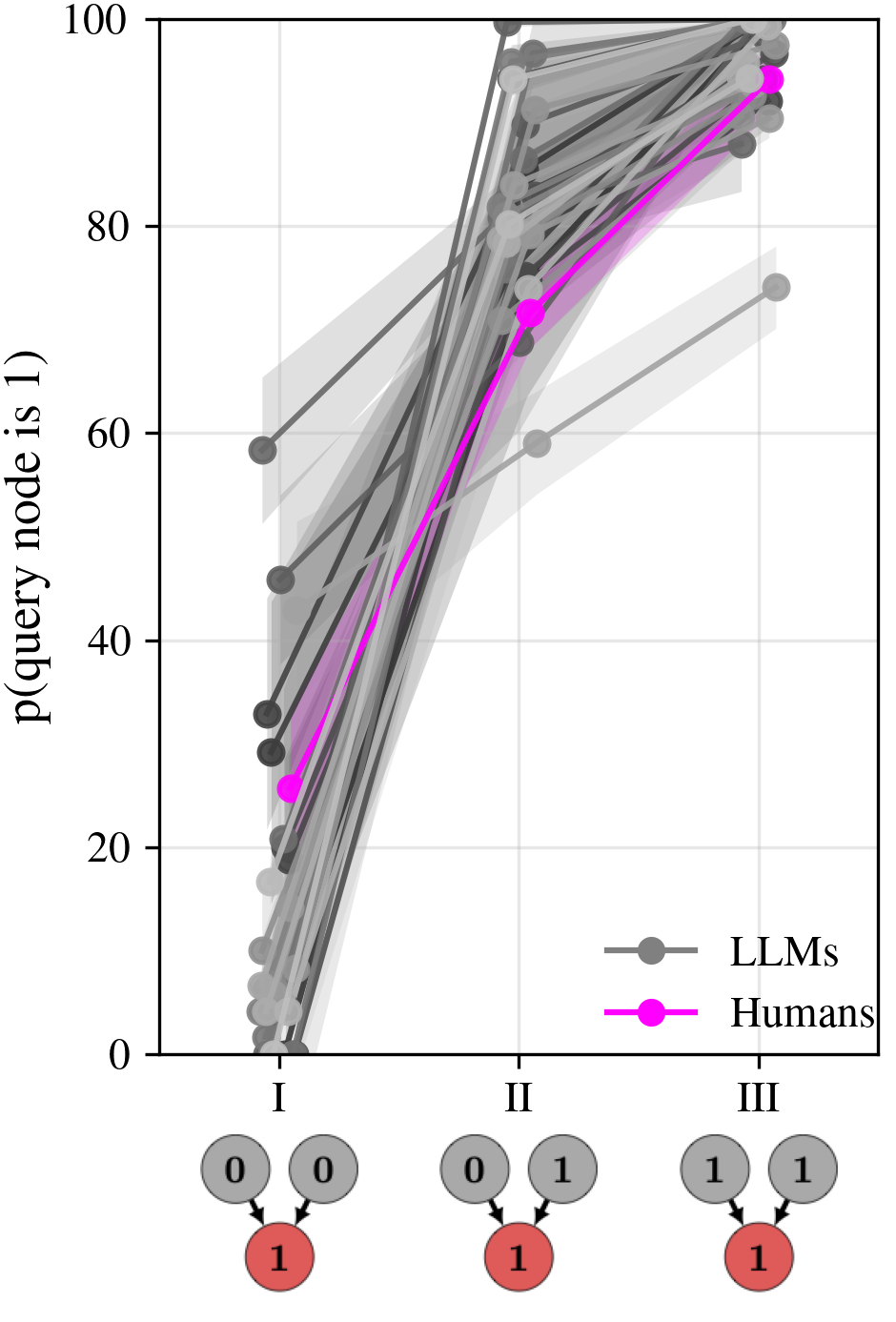}
        \caption{\small }
        \label{fig:sub_pred_inference}
    \end{subfigure}\hfill
    \begin{subfigure}[b]{0.32\linewidth}
        \centering
        \includegraphics[width=\linewidth]{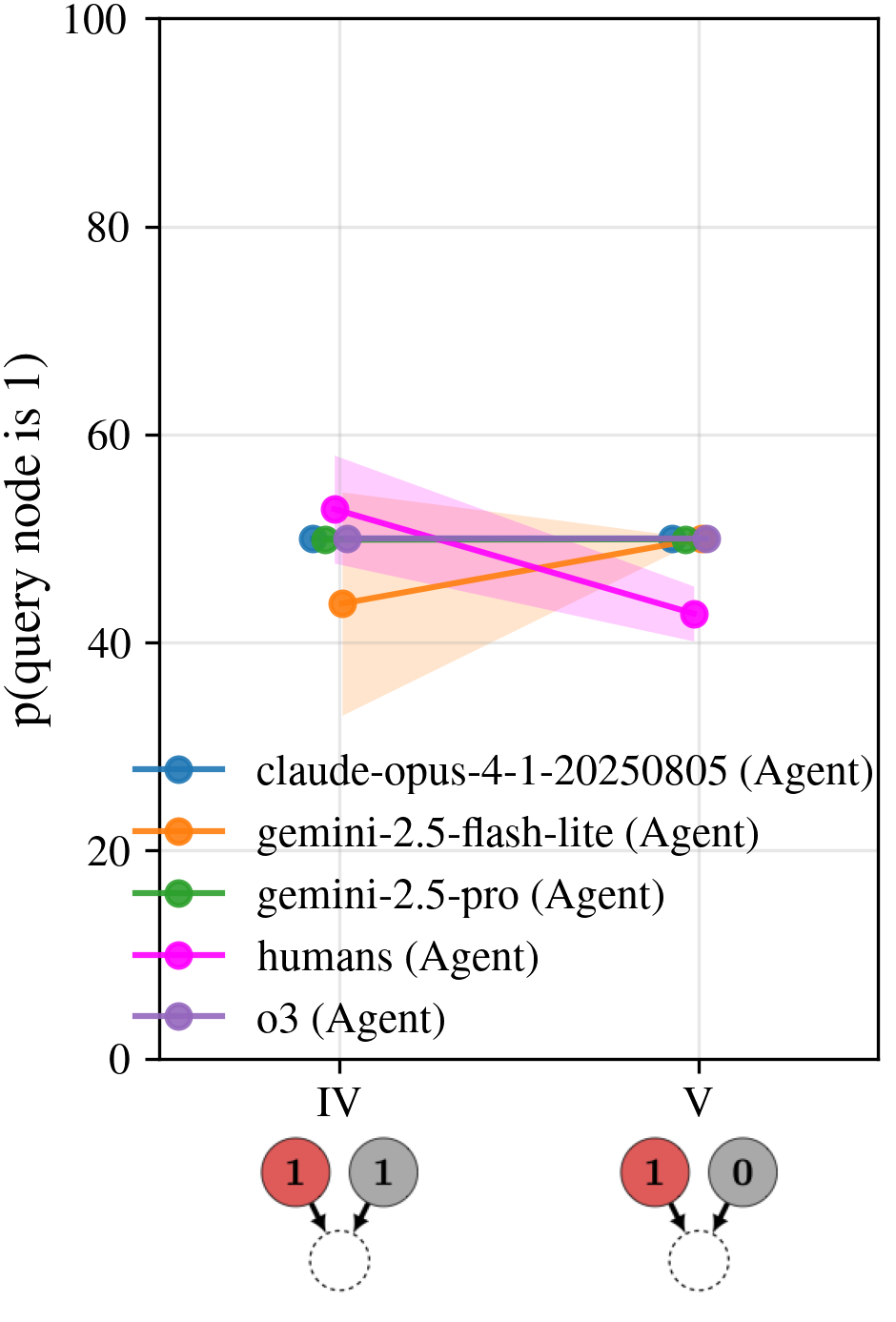}
        \caption{\small }
        \label{fig:bias_mv}
    \end{subfigure}\hfill
    \begin{subfigure}[b]{0.33\linewidth}
        \centering
        \includegraphics[width=\linewidth]{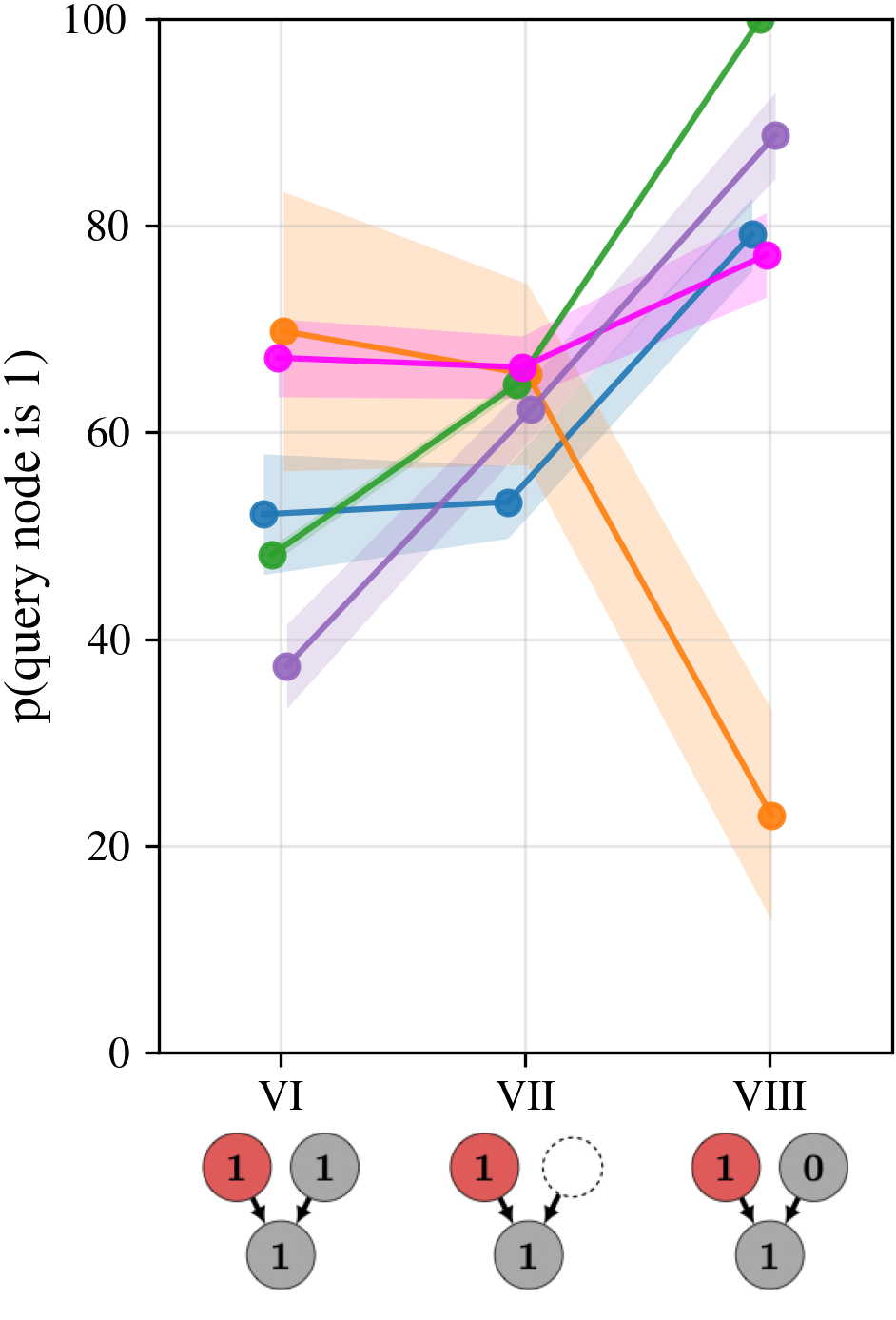}
        \caption{\small }
        \label{fig:bias_ea}
    \end{subfigure}
   \caption{
    \small Like humans, LLMs judge the effect as more likely in the presence of more causes (\Cref{fig:sub_pred_inference}). Unlike humans, most LLMs (a) do not reproduce the Markov violation bias and instead respect the independence of causes (\Cref{fig:bias_mv}) and (b) show strong explaining away where humans show only a weak effect (\Cref{fig:bias_ea}).
    }
    \label{fig:tasks}
\end{figure}

\textbf{Both LLMs and humans judge the effect as more probable the more causes are present.}
We first look at the three predictive inference tasks in our dataset where given one of the causes (known), the probability of the effect (to be predicted) is judged.
\Cref{fig:sub_pred_inference} shows the results for these tasks, where we find that both LLMs (shades of gray) and humans (pink) follow the same pattern: the more causes are present, the higher the judged probability of the effect, which we take as evidence for agents understanding the causal mechanism and providing sensible judgments.

\begin{wrapfigure}[30]{r}{0.6\textwidth}
  \centering
    \includegraphics[width=\linewidth,trim=0 0.08cm 0 0.05cm, clip]{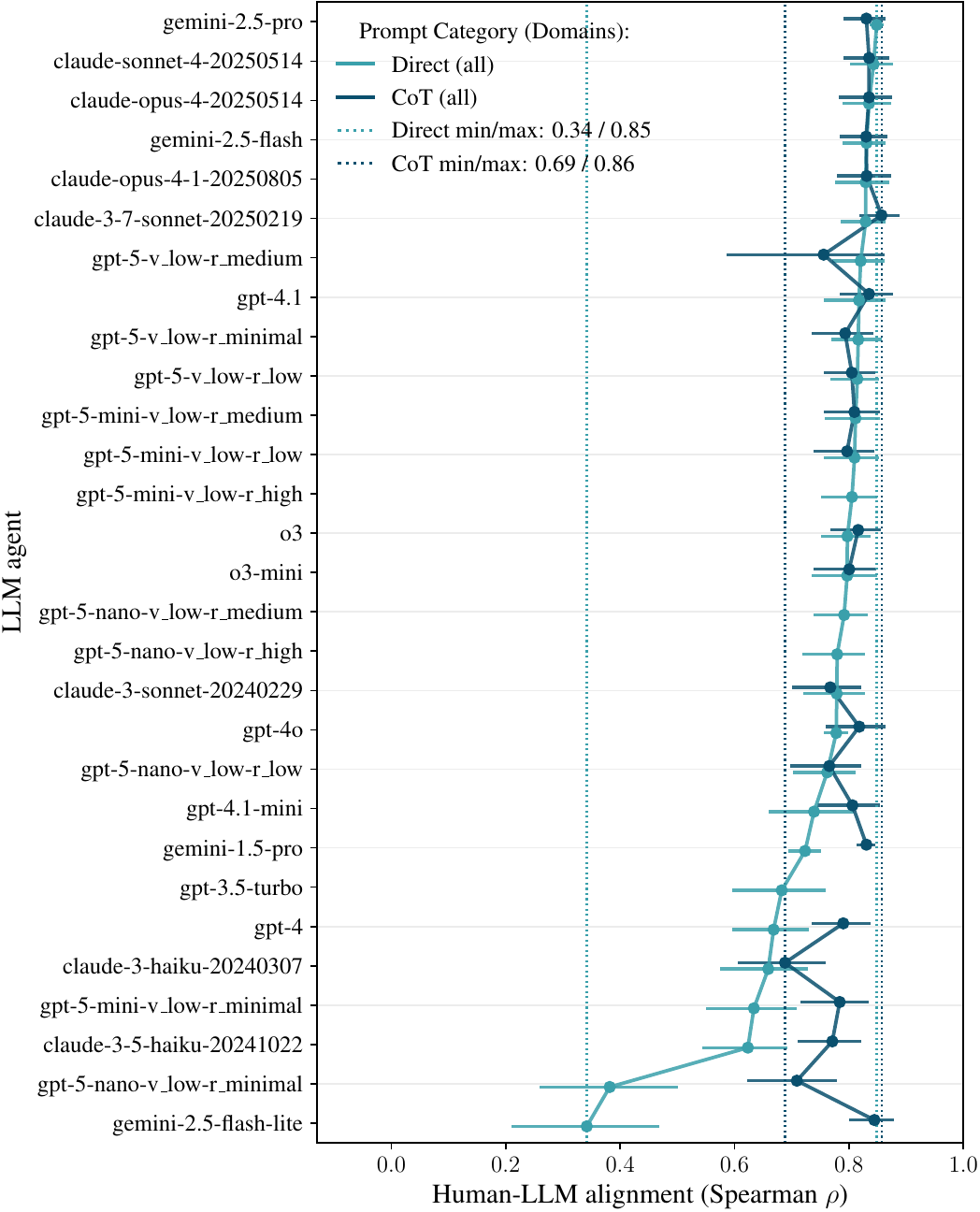}
  \caption{
\small
 Human--LLM alignment: CoT \cotcircle{}   boosts alignment.
}
  \label{fig:human-llm-alignment}
\end{wrapfigure}

\textbf{Alignment Increases under Chain-of-Thought Prompting}
\Cref{fig:human-llm-alignment} shows human--LLM alignment measured via Spearman correlation ($\rho$) between the probability judgments of humans and each LLM  (95\% CIs via 2000 bootstrap resamples), under both Direct and CoT prompting.
CoT prompting generally increases alignment for less aligned LLMs under the direct prompting condition \numericcircle.

\subsection{Q2: Can small interpretable (cognitively-grounded) models capture / predict LLM reasoning?}
\label{sec:results_normativity_cbn}

\vspace{1em}
\begin{tldr*}
  LLM causal judgments are compressible by  small interpretable causal Bayes net (CBN)models.
  Predicting causal judgements on held out tasks during fitting, we find that \textit{chain-of-thought (CoT)} increases  the alignment of CBN and LLM probability judgments, especially pronounced in noisy conditions.
\end{tldr*}

\textbf{Rationale.}
All tasks share the same underlying collider structure ($C_1 \!\rightarrow \!E\! \leftarrow \!C_2$).
Since RW17 does not specify causal strengths or base rates, there is no unique ground-truth probability judgment to score against \cite{rehder2017failures}.
We therefore evaluate \textit{causal-model consistency}:
whether an agent's probability judgments are well predicted by a compact causal model with the causal graph   structure that formalizes the tasks.

\textbf{Causal Bayes nets as small interpretable causal  judgment  baseline.}
For each agent, prompt condition, and 4 experimental conditions, we fit a causal Bayesian network (CBN) with a leaky noisy-OR parameterization \cite{cheng1997causalpower},
 pooling judgments across all 11 collider tasks $t$ and domains,  where the 11 tasks refer to different conditional-probability queries on the collider graph (see \Cref{fig:task-introx},  \Cref{fig:tasks}).
For binary causes $C_1,C_2 \!\in \!\{0,1\}$, effect $E$, parameters $\theta$ background leak $b\!\in[0,1]$ and causal strengths $m_1,m_2\in[0,1]$, the conditional probability is
$
\Pr(E\!{=}\!1 \!\mid\! C_1, C_2)
\!= \!1\! - \!(1-b)\,(1 \!- \!m_1 )^{C_1}\,(1\! - \!m_2 )^{C_2}.
$
This  captures that each active cause independently increases the probability of the effect, while the leak
$b$ accounts for background activations in the absence of any
active causes.
We fit parameters by minimizing squared error between model predictions $\hat{y}_t(\theta)$ and an agent's normalized probability judgments $y_t$ over tasks $t$.


\begin{figure}[htbp]
 \includegraphics[width=\linewidth,  trim=0 0.48cm 0 1.64cm, clip]{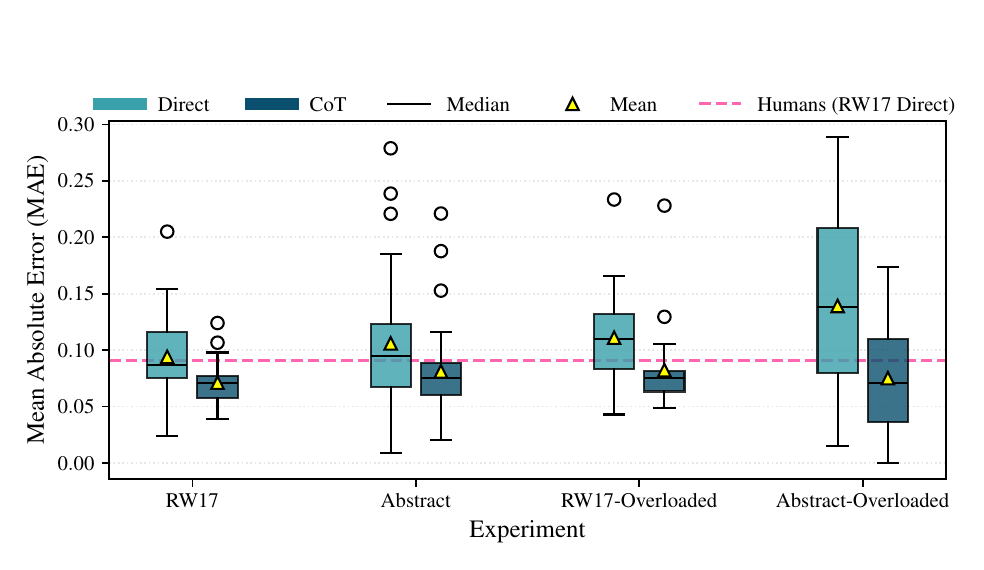}
  \caption{\small
  LLMs' probability judgments are well captured by an
interpretable causal model, indicated by good causal Bayes net fits; CoT improves fit.
  }
  \label{fig:cbn-error-mae}
\end{figure}

\begin{figure}[htbp!]
    \vspace{-1em}
    \centering
     \includegraphics[width=.95\linewidth,trim=0 0.5cm 0 1.5cm, clip]{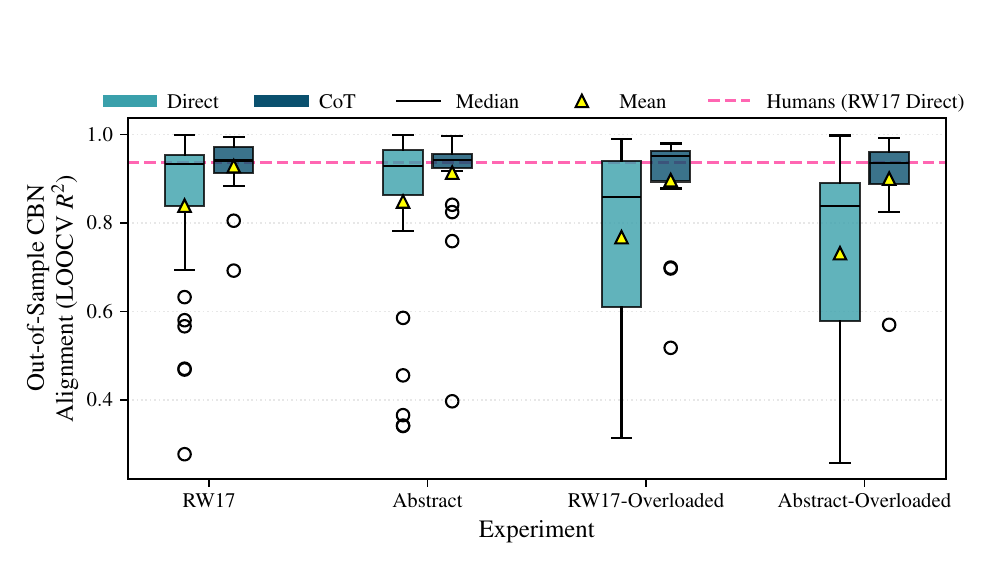}
    \caption{\small Out-of-sample CBN generalization (LOOCV $R^2$); higher is better.}
    \label{fig:loocv-r2-boxes}
\end{figure}
\textbf{LLMs' probability judgments are well captured by an interpretable causal model}
\Cref{fig:cbn-error-mae} reports the mean absolute error (MAE) $\in \![0,1]$ (lower is better) for the fitted CBNs.
Across experiments, most agents have a small MAE and are hence well-described by a single  CBN. CoT \cotcircle{}  prompting reduces MAE and dispersion relative to direct \numericcircle{}   prompting.
Overloaded prompts increase error, especially in the abstract setting. This indicates  that both the direct prompting style \numericcircle{}   and  irrelevant content  has agents apply more varied causal reasoning strategies.

\textbf{CoT improves out-of-sample CBN generalization.}
To test whether an agent's judgments are captured by a \textit{single} fitted CBN in a way that generalizes \textit{beyond} the tasks used for fitting (rather than reflecting task-specific idiosyncrasies), we compute \emph{out-of-sample CBN alignment} as task-level leave-one-out cross-validation (LOOCV) $R^2$ (higher is better; $R^2\!\in\!(-\infty,\!1]$).
Concretely, we fit the CBN on 10 of 11 collider tasks,
 predict the held-out task, pool held-out predictions across folds, and compute $R^2$ on held-out judgments (\Cref{fig:loocv-r2-boxes}).
CoT increases LOOCV $R^2$ and tightens dispersion across agents, with the largest gains under overloaded prompts, indicating that CoT improves the cross-task generalization of CBN-consistent causal reasoning under distraction.
Several models approach or exceed the \emph{human benchmark} (pink dashed line) and more so under CoT \cotcircle{}.

%


\subsection{Q3: How much do agents attribute to external, unmentioned factors?}
\vspace{1em}
\begin{tldr*}
Most LLMs act as stricter rule-followers than humans, showing high reliance on the stated causal links and low account for latent, unmentioned factors.
Chain-of-Thought can shift models toward tighter causal rule-following in some cases, with mixed effects overall.
\end{tldr*}

\textbf{Rule-Following\ vs.\ Background Attribution:}
When agents interpret an explicitly stated causal model, they can differ in how much they treat the stated causes as \textit{sufficient} versus how much they attribute outcomes to \textit{external, unmentioned factors}.
We operationalize this spectrum in a noisy-OR model via \textit{Background-Adjusted Causal Strength} (BACS) $\!\in\![-1,1]$:
$
    \mathrm{BACS}_{\text{agent}}\! =\! \overline{m}_{\text{agent}} \!- \!b_{\text{agent}}
    $, where $\overline{m}_{\text{agent}}$ is the mean of the causal strength parameters $m_{1,2}$ (how sufficient causes are perceived to be, $m\!=\!1$ means completely sufficient) and $b_{\text{agent}}$ is the leak/background parameter (baseline prevalence of the effect in the \textit{absence} of the stated causes, capturing the influence of latent or unmentioned factors  on the causal system).
High $\mathrm{BACS}\!\to\!1$ indicates \textit{high causal rule-fidelity}: agents largely treat the provided causal relationships as sufficient (high $m$, low $b$).
Low $\mathrm{BACS}\!\to \!-\!1$  reflects  stronger \textit{latent-factor accommodation} (elevated $b\!>\!0$) and causes not being sufficient ($m \!< \!1$).
 We therefore treat BACS as a summary index of reliance on the stated rule versus accounting for non-specified latent factors.

\begin{figure}[htbp]
 \includegraphics[width=\linewidth,trim=0 0.5cm 0 0.5cm, clip]{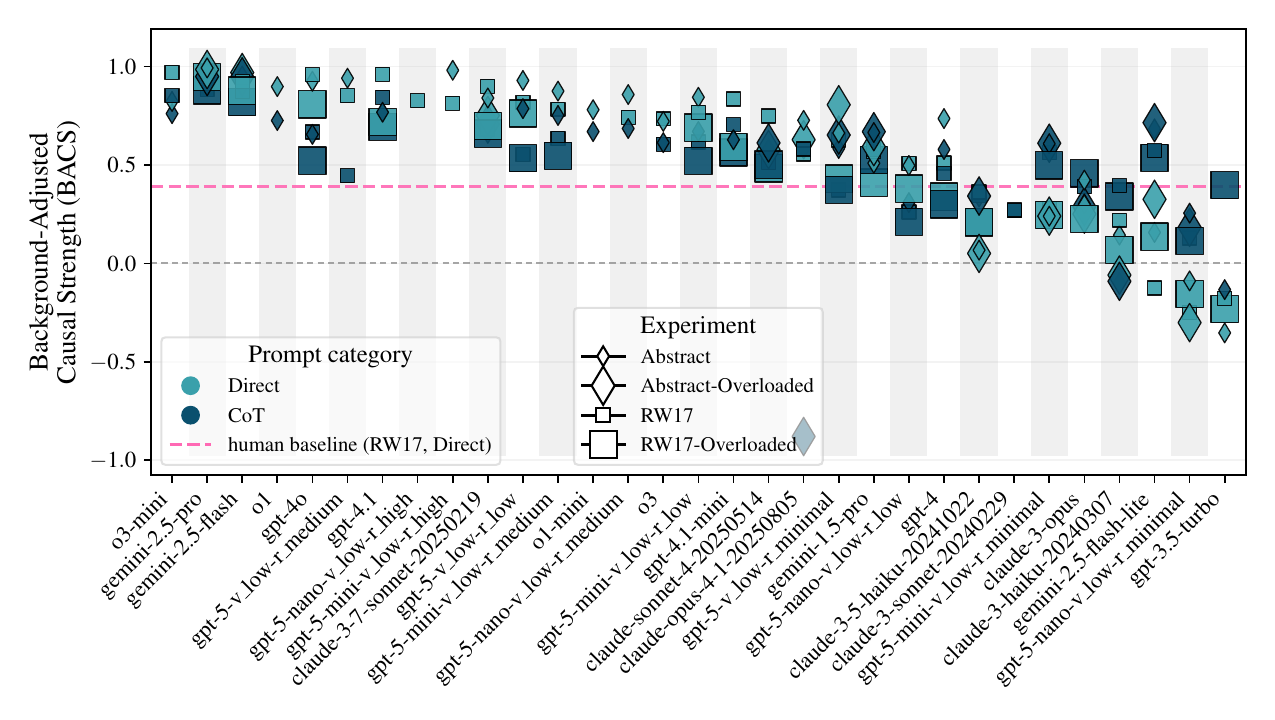}
  \caption[By release date: (BACS)]{\small Background-Adjusted Causal Strength (BACS) per experiment and prompt category by LLM. Most LLMs have much higher rule-fidelity than humans.  Higher BACS indicates tighter causal rule-following (stronger reliance on stated links relative to background leak), whereas lower BACS indicates greater background attribution and weaker adherence to the stated causal strengths.}
  \label{fig:by_release_date_det}
\end{figure}
\textbf{Most LLMs show higher causal rule-fidelity than humans}
\Cref{fig:by_release_date_det} shows Background-Adjusted Causal Strength (BACS) per experiment and prompt category by LLM.
Most agents lie above the human benchmark (pink line) across experimental conditions marked by shape and size of scatters in \Cref{fig:by_release_date_det} indicating that LLMs typically follow the stated causal rule more tightly than humans, whereas humans
 seem to be more prone to account for latent factors and treat the stated causes as insufficient.
Some agents (e.g., \texttt{gemini-2.5-pro} and \texttt{gemini-2.5-flash}) approach near-maximal rule-fidelity ($\mathrm{BACS}{\approx}1.0$) and are agnostic to experimental conditions as indicated by tight clusters. A smaller set of agents exhibits lower BACS than humans in the RW17-Direct condition (right side in \Cref{fig:by_release_date_det}), treating the stated causal links as comparatively weak and  expecting the effect to occur even in the absence of  the stated causes. With CoT, these LLMs approach or slightly exceed the human baseline, whereas already highly rule-following models remain largely unaffected by prompting strategy or even become slightly less rule following with CoT.



\subsection{Q4: Are LLMs biased like humans?}

\vspace{1em}
\begin{tldr*}
Humans often show two biases in collider graphs: \emph{weak} explaining away (EA) and \emph{associative} dependence between causes (Markov violations (MV)).
Across experiments, most LLMs show stronger explaining away than humans and are largely Markov-compliant; a minority exhibit human-like associative Markov violations.
This suggests that most LLMs \emph{do not} mirror the characteristic human collider biases (weak EA + MV).
\end{tldr*}

\begin{wrapfigure}[16]{r}{0.7\textwidth}
    \vspace{-1em}
    \centering
     \includegraphics[width=.95\linewidth,trim=0 0.5cm 0 1.5cm, clip]{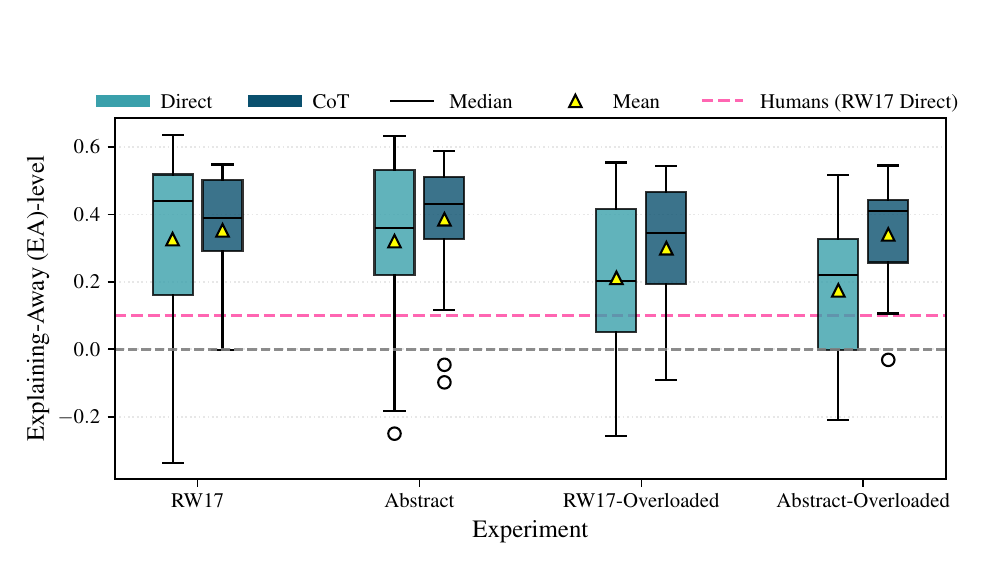}
    \caption{\small LLMs exhibit stronger explaining away (EA) than humans, indicated by EA-level $\gg 0$; Chain-of-Thought prompting enhances this effect in many models.}
    \label{fig:ea}
 \end{wrapfigure}

\textbf{Bias signatures in collider reasoning: no / weak explaining away and Markov violations}
In a collider graph, two qualitative signatures sharply distinguish normative causal inference from common human shortcuts.
First, \emph{explaining away} (EA)
occurs when  evidence for one cause reduces the belief in the alternative cause in the presence of the effect. We quantify the level of explaining away (EA) directly from agents' raw normalized probability judgments
$
\mathrm{EA} \!=\! \Pr(C_1{=}1 \!\mid\! E\!{=}\!1,C_2\!{=}\!0) \!-\! \Pr(C_1\!{=}\!1 \!\mid\! E\!{=}\!1,C_2\!{=}\!1) 
$, where $EA \!\leq\! 0$ represents absense of explaining away and the greater EA is, the more explaining away occurs.
Qualitatively, explaining away corresponds to a positive  slope in \Cref{fig:bias_ea}.
Second, \emph{Markov violation} captures the violation of the   independence of causes in the absence of evidence about the effect:
$
\mathrm{MV} \!=\! \big|\Pr(C_1{=}1 \!\mid\! C_2{=}1) \!-\! \Pr(C_1{=}1 \!\mid\! C_2{=}0)\big|,
\quad \text{with Markov compliance when } \mathrm{MV}\!\approx \!0.
$
Qualitatively  Markov compliance corresponds to a flat line  in \Cref{fig:bias_mv}.
Humans frequently deviate on both:
they often show \textit{too little or no} explaining away
 \cite{fernbach2013cognitive, rottman2014reasoning} meaning evidence for one cause fails to reduce belief in an alternative cause.
Humans also have been shown to  \textit{violate} cause-independence (Markov violation), meaning their judgment of one cause is   influenced by the presence or absence of an alternative cause \cite{ali2011mental, mayrhofer2016sufficiency, park2013mechanistic, rehder2017failures}.
Whether LLMs show similar human biases is an open question we address in this work.
\begin{wrapfigure}[14]{r}{0.7\textwidth}
    \vspace{-1em}
    \centering
     \includegraphics[width=.95\linewidth,trim=0 0.5cm 0 1.5cm, clip]{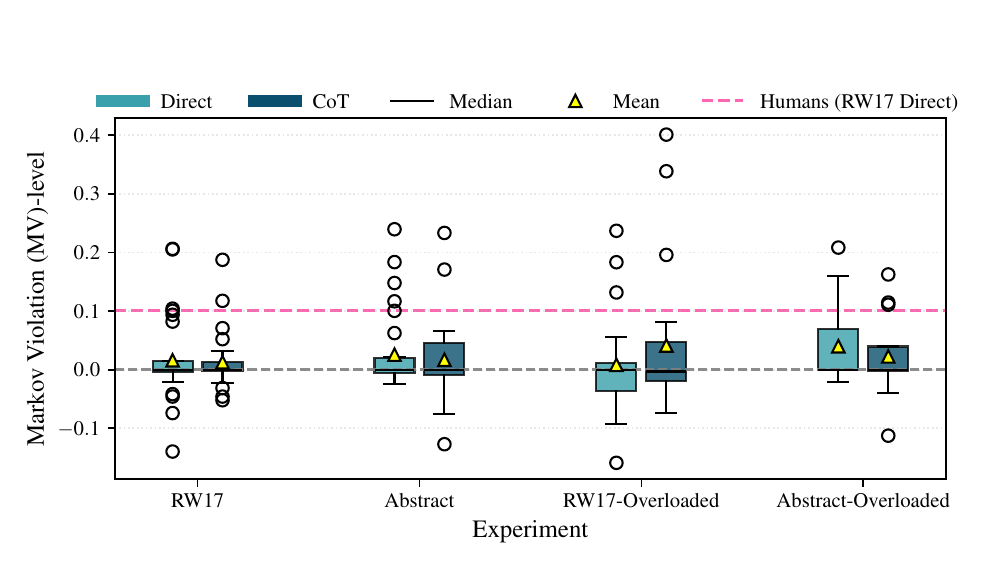}
    \caption{\small  LLMs mostly don't exhibit Markov violation (MV), indicated by MV-level $\!\approx \!0$, while humans do.}
    \label{fig:mv}
\end{wrapfigure}

\textbf{LLMs rarely reproduce  human collider biases}
Across agents, explaining away is common and typically much stronger than in humans indicated by the mean and median EA-levels for LLMs in \Cref{fig:ea} falling above the human baseline (pink, \(\mathrm{EA}\!\approx \!0.1\)).
Under overload, many LLMs show reduced EA relative to baseline, with CoT largely restoring EA toward baseline levels.


\textbf{Most LLMs are Markov compliant; a minority show associative bias like humans.}
For the  Markov condition, the modal pattern is \textit{compliance}: most LLMs cluster tightly around \(\mathrm{MV}\!=0\!\)  in \Cref{fig:mv}, indicating that one cause does not spuriously change belief in the other whereas for humans  this is true.
Under overload, Markov violations occur more frequently, albeit still less than in humans.


\begin{figure*}[htbp!]
  \centering
    \begin{subfigure}[b]{\textwidth}
    \centering
    \includegraphics[width=\linewidth]{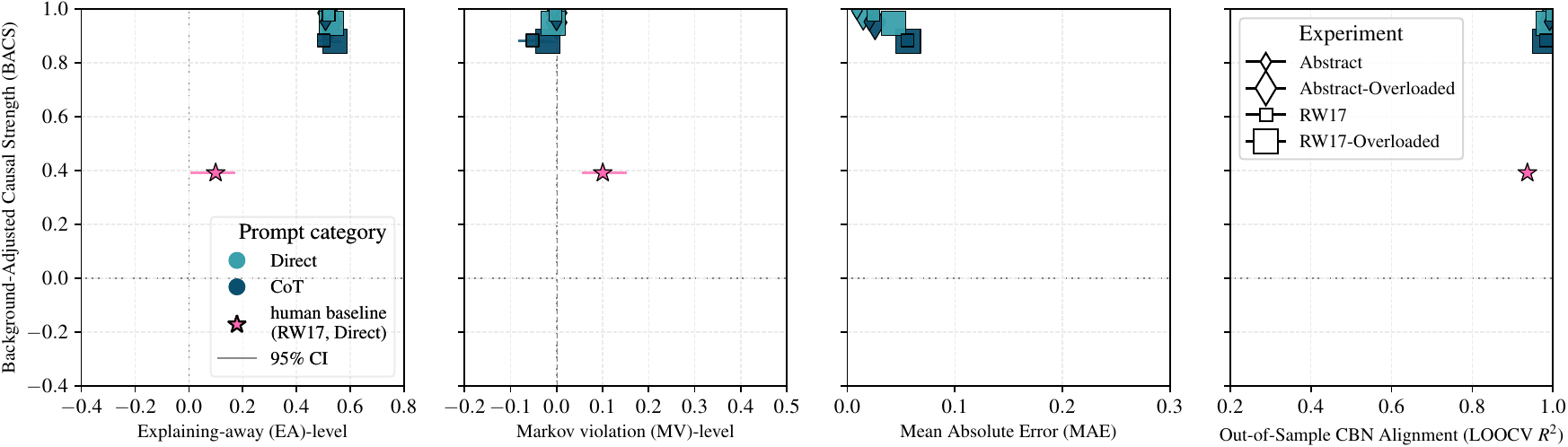}   \caption{\small \emph{Gemini-2.5-pro} reasons most robustly indicated by tight clusters among all 8 experimental conditions, showing content and prompting invariance.}
    \label{fig:bacs_gemini-2.5-pro}
  \end{subfigure}%
  \vfill
  \begin{subfigure}[b]{\textwidth}
    \centering
 \includegraphics[width=\linewidth]{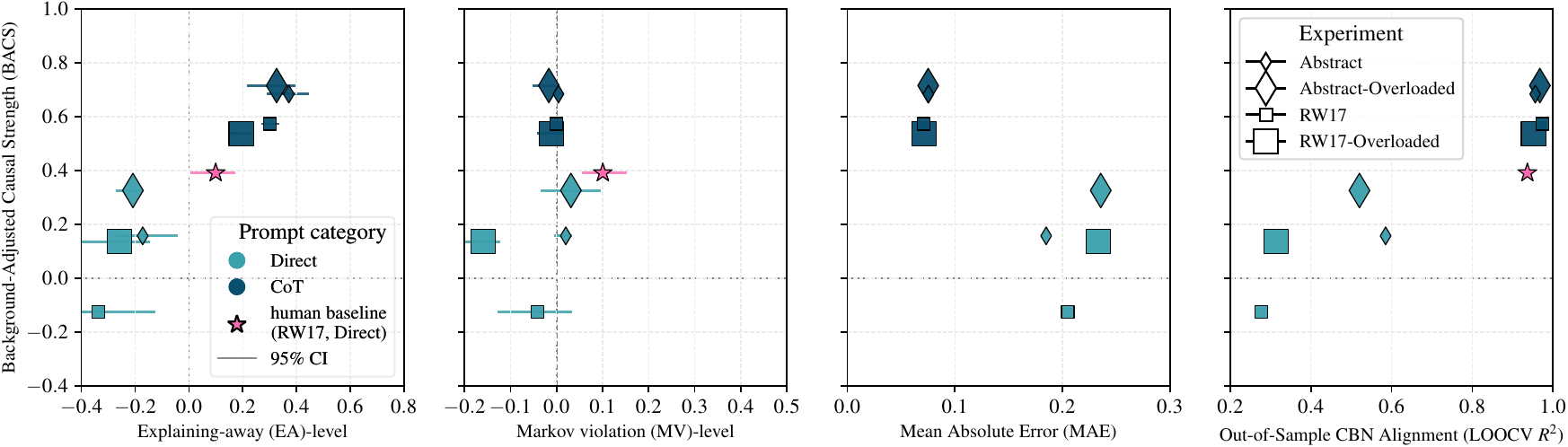}    \caption{\small \emph{Gemini-2.5-flash-lite} shows clear prompting effects indicated by color-separated clusters  (Direct  \numericcircle{}, CoT \cotcircle{}) with CoT \cotcircle{} increasing content invariance drastically.}
    \label{fig:bacs_gemini-2.5-flash-lite}
  \end{subfigure}%
  \caption[Background-Adjusted Causal Strength (BACS) levels]
  {\small Causal reasoning robustness across experimental conditions for metrics discussed in this work. Tight clustering indicates greater robustness to content manipulations and prompt variations.}
  \label{fig:robustness_main}
 \end{figure*}

\subsection{Q5: Do LLMs Reason Robustly Under Content Manipulations?}
\label{sec:results-robust}

\vspace{1em}
\begin{tldr*}
Robustness remains highly model-dependent; Gemini-2.5-pro is near-invariant across all 8 experimental conditions. CoT typically increases robustness.
\end{tldr*}

We evaluate whether agents preserve their causal-reasoning signatures under all 8 experimental conditions involving (i) content abstraction (RW17$\to$Abstract), (ii) noise injection/overload, and (iii) prompting strategy (Direct \numericcircle{} vs.\ CoT \cotcircle{}).
\Cref{fig:robustness_main} summarizes robustness for a select number of LLMs at a glance across metrics considered in this work involving Background-Adjusted Causal Strength (BACS), Causal-Bayes-Net Alignment (LOOCV $R^2$), explaining away level (EA), and Markov Violation level (MV): each agent forms a cluster (robust) across all eight experimental conditions  or separates (not robust).

\textbf{Robustness is strongly model-dependent and CoT improves it for many models.}
\textit{Gemini-2.5-pro} (\Cref{fig:bacs_gemini-2.5-pro}) shows tight clustering across all 8 conditions (high BACS, high $R^2$, strong EA, MV$\approx 0$), indicating near-invariant collider reasoning under all 8 manipulations.
In contrast, \textit{Gemini-2.5-flash-lite} (\Cref{fig:bacs_gemini-2.5-flash-lite}) shows clear prompting effects:
For Direct \numericcircle{} prompting, clusters are more dispersed, while for CoT \cotcircle{} prompting, clusters are much tighter signaling higher content invariance when doing causal reasoning.
CoT often increasing robustness by reducing the cluster size like for \textit{Gemini-2.5-flash-lite} (\Cref{fig:bacs_gemini-2.5-flash-lite}) was e observed for several other models as well.
CoT often increased robustness reducing the cluster size like for \textit{Gemini-2.5-flash-lite} (\Cref{fig:bacs_gemini-2.5-flash-lite}).


\section{Discussion}
Our results suggest that many contemporary LLMs apply causal reasoning in collider graphs in a comparatively rule-like manner.
Humans, by contrast, allocate more mass to unmentioned background factors (lower Background-adjusted causal strength; \Cref{fig:by_release_date_det}), consistent with more open-world interpretations of the prompt.
Yet LLMs’ strong rule-following reasoning style
risks failure in real-world settings where uncertainty is intrinsic,
underscoring the need to better characterize LLM reasoning
strategies to guide their safe and effective application.
 Despite being trained largely on human-generated text, most LLMs do not reproduce human collider biases: they show strong explaining away (EA) and are largely Markov-compliant, whereas humans exhibit low EA and frequent Markov violations. Robustness is strongly model-dependent and shaped by prompting: abstraction and overload shift causal signatures for many models, while CoT often restores more stable behavior; newer, larger models (e.g., Gemini-2.5-pro) are near-invariant across content manipulations, whereas smaller/older models are markedly more sensitive. Our results further show that both human and machine causal judgments can be predicted by—and compressed into—small interpretable causal Bayes nets. More broadly, this illustrates how Bayesian models can provide computational-level targets that help interpret and diagnose the behavior of multi-billion paramter  neural networks,  exemplifying that  Bayesian and neural network models are complementary in understanding intelligence \citep{griffiths2024bayes}.

A key limitation is our focus on the common-effect (collider) graph. We chose this structure because colliders elicit well-studied signatures such as explaining away and because decades of work shows systematic human biases in precisely these settings; however, this limits the generality of the present conclusions. Additionally, our prompts do not explicitly specify whether unmentioned causes should be ignored or treated as plausible background factors, which may differentially affect humans and models and could partly explain the observed closed-world versus open-world gap. More broadly, the degree to which these laboratory-style tasks capture causal reasoning under real-world uncertainty remains an open question. Lastly, we don't have human responses for the overloaded and abstract conditions.

\textbf{Future work} should address the limitations above.
Furthermore, human-LLM tandems could be explored to better understand how these models might complement human reasoning.


\setlength{\bibleftmargin}{.125in}
\setlength{\bibindent}{-\bibleftmargin}

\bibliography{bibliography.bib}



\end{document}